\documentclass[conference]{IEEEtran}
\IEEEoverridecommandlockouts
\usepackage{cite}
\usepackage{amsmath,amssymb,amsfonts}
\usepackage{algorithmic}
\usepackage{graphicx}
\usepackage{comment}
\usepackage{textcomp}
\usepackage{xcolor}
\usepackage{booktabs}
\usepackage{amssymb}
\usepackage{rotating} 
\usepackage[utf8]{inputenc}
\usepackage{url}
\usepackage{subcaption}
\def\BibTeX{{\rm B\kern-.05em{\sc i\kern-.025em b}\kern-.08em
    T\kern-.1667em\lower.7ex\hbox{E}\kern-.125emX}}
\begin{document}

\title{SCONE: A Practical, Constraint-Aware Plug-in for Latent Encoding in Learned DNA Storage
\\}

\author{
\IEEEauthorblockN{
Cihan Ruan\IEEEauthorrefmark{1}\textsuperscript{\textdagger},
Lebin Zhou\IEEEauthorrefmark{1}\textsuperscript{\textdagger},
Rongduo Han\IEEEauthorrefmark{3},
Linyi Han\IEEEauthorrefmark{1},
Bingqing Zhao\IEEEauthorrefmark{4}, 
Chenchen Zhu \IEEEauthorrefmark{4}\\
Wei Jiang\IEEEauthorrefmark{5},
Wei Wang \IEEEauthorrefmark{5}
and Nam Ling\IEEEauthorrefmark{1}{\textsuperscript{$\blacklozenge$}}
}
\IEEEauthorblockA{
\IEEEauthorrefmark{1}Department of Computer Science and Engineering, Santa Clara University, Santa Clara, CA, USA\\
\IEEEauthorrefmark{3}College of Software, Nankai University, Tianjin, China\\
\IEEEauthorrefmark{4}Department of Genetics, Stanford School of Medicine, Stanford University, Palo Alto, CA, USA\\
\IEEEauthorrefmark{5}Futurewei Technologies Inc., Santa Clara, CA, USA\\
Email: luciacihanruan@gmail.com, lzhou@scu.edu, hrd12910@gmail.com, linyihan25@gmail.com\\ {\{bqzhao, czhu5\}@stanford.edu}, {\{wjiang, rickweiwang\}@futurewei.com}, nling@scu.edu
}
}

\maketitle

\renewcommand{\thefootnote}{\fnsymbol{footnote}}
\footnotetext{\textsuperscript{\textdagger}These authors contributed equally to this work.}
\footnotetext{\textsuperscript{$\blacklozenge$}Corresponding author.}
\renewcommand{\thefootnote}{\arabic{footnote}}

\begin{abstract}
DNA storage has matured from concept to practical stage, yet its integration with neural compression pipelines remains inefficient. Early DNA encoders applied redundancy-heavy constraint layers atop raw binary data—workable but primitive. Recent neural codecs compress data into learned latent representations with rich statistical structure, yet still convert these latents to DNA via naive binary-to-quaternary transcoding, discarding the entropy model's optimization. This mismatch undermines compression efficiency and complicates the encoding stack. A plug-in module that collapses latent compression and DNA encoding into a single step. SCONE performs quaternary arithmetic coding directly on the latent space in DNA bases. Its Constraint-Aware Adaptive Coding module dynamically steers the entropy encoder's learned probability distribution to enforce biochemical constraints— Guanine-Cytosine (GC) balance and homopolymer suppression—deterministically during encoding, eliminating post-hoc correction. The design preserves full reversibility and exploits the hyperprior model's learned priors without modification. Experiments show SCONE achieves near-perfect constraint satisfaction with negligible computational overhead ($<2\%$ latency), establishing a latent-agnostic interface for end-to-end DNA-compatible learned codecs. 

\textbf{Code:} \url{https://github.com/cruan1991/SCONE-DNA}
\end{abstract}


\section{Introduction}

DNA has matured from a storage medium into an intelligent computational substrate\cite{church2012next, bornholt2016dna, goldman2013towards, grass2015robust} DNA-based neural networks now demonstrate pattern classification through supervised learning~\cite{cherry2025supervised} and sustained computation via heat-recharging~\cite{song2025heat}, while bio-chips~\cite{ gupta2024artificial} enable system integration~\cite{ricouvier2024largescale, yang2024spatial}. This convergence of storage, learning, and in-memory processing~\cite{zhang2024high} establishes DNA as a foundation for embedded bio-AI systems\cite{beyond_silicon_dna}.

There is a bottleneck that persists: encoding.  
Most existing pipelines remain bit-centric—compressing data in binary and only later remapping it into the quaternary alphabet $\{A,T,G,C\}$.  
While approaches such as DNA Fountain~\cite{erlich2017dna} and HEDGES~\cite{press2020hedges} achieve robustness via parity and redundancy, they treat biochemical constraints as an afterthought, resulting in inefficiency and broken differentiability when coupled with neural codecs~\cite{ruan2025hybridflow,su2025robust,11210168}.

To address this gap, we propose \textbf{SCONE} (\textbf{S}implified \textbf{C}onstraint-aware \textbf{ON}-network \textbf{E}ncoder), a plug-in module that replaces binary entropy coding with native quaternary arithmetic coding.  
SCONE operates directly on the nucleotide alphabet under a finite-state constraint regularization mechanism, satisfying Guanine-Cytosine (GC) balance and homopolymer limits during probability modeling. While prior neural-based systems utilized image-specific redundancy for robust decoding\cite{ruan2024robust, ruan2025dsi}, SCONE focuses on native constraint satisfaction to maximize encoding density. It requires no post-hoc correction or redundancy injection, introduces less than 2\% runtime overhead, and is compatible with any learned compression framework.

Our contributions include a reversible, codec-agnostic quaternary interface that performs arithmetic coding directly on DNA alphabets, and a deterministic constraint-aware probability guidance mechanism that eliminates post-hoc correction layers. Experimental validation demonstrates more than 99\% constraint satisfaction with negligible rate and latency overhead, establishing practical viability for integrated DNA-neural systems.

\begin{figure*}[htp]
    \centering
    \includegraphics[width=13.5cm]{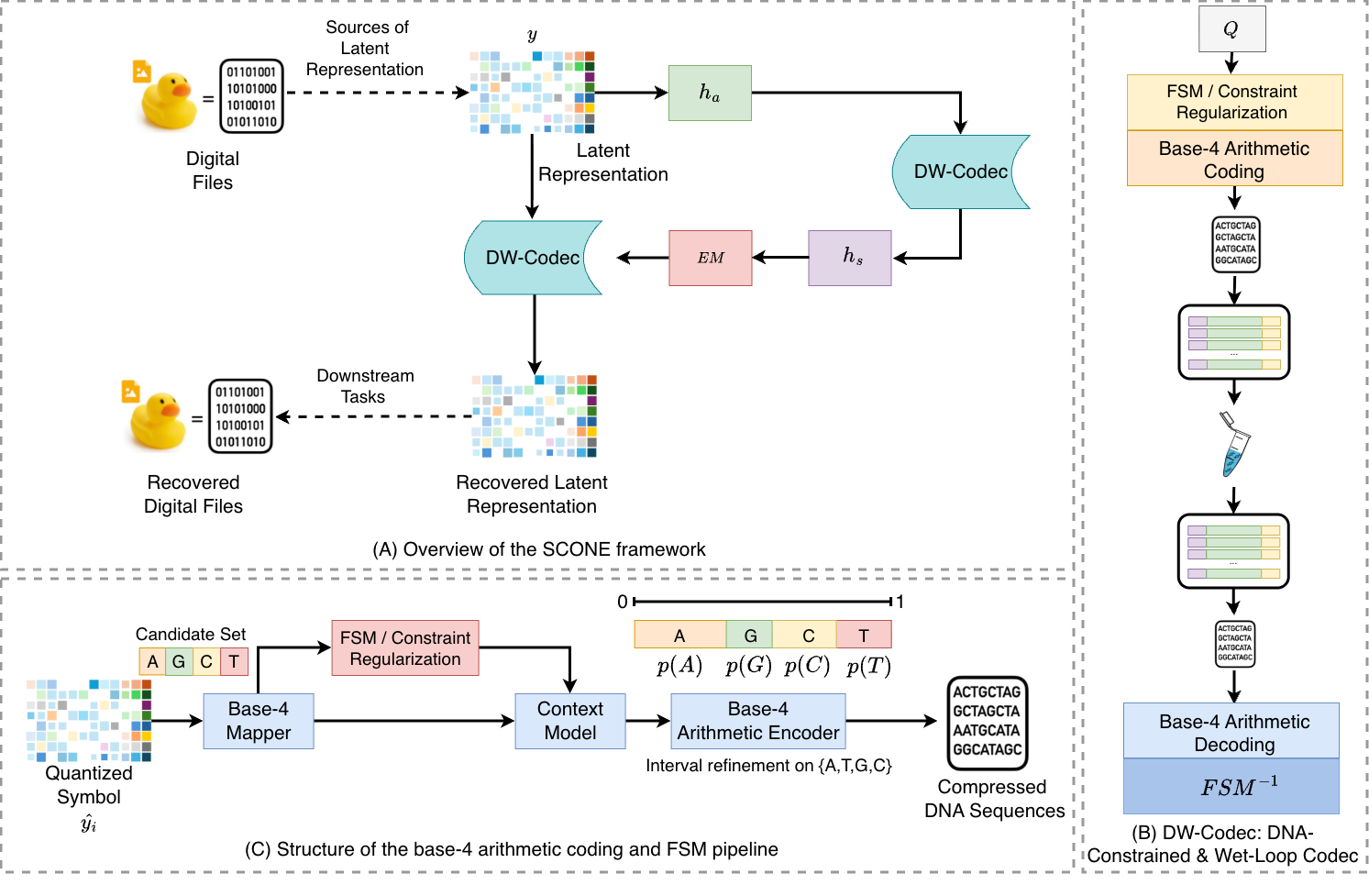}
    \caption{
        \textbf{SCONE framework overview.} 
        \textbf{(A)} The SCONE module is designed as a plug-in neural encoder that integrates seamlessly into existing machine learning pipelines. It accepts latent representations $y$ from arbitrary upstream models (e.g., autoencoders, VAEs, image compressors), and transforms them into DNA-compatible representations. After biochemical storage and decoding, the reconstructed latent can be passed into downstream tasks such as lossless recovery of latent tokens or semantic retrieval—demonstrating SCONE’s adaptability to diverse latent spaces.
        \textbf{(B)} \textbf{DW-Codec} (DNA-Constrained \& Wet-loop Codec): a codec stack that translates quantized representations into synthesizable DNA strands via base-4 arithmetic coding and constraint filtering. The decoder path reverses this transformation after sequencing.
        \textbf{(C)} FSM-regularized base-4 arithmetic encoding: quantized latent symbols are mapped to ATGC bases using a context model. Finite-State Machines (FSMs) are used to filter illegal patterns (e.g., homopolymers or unbalanced GC-content) before base-4 interval encoding.
    }
    \label{fig:workflow}

\end{figure*}

\section{Related Work}

\subsection{DNA Encoding Evolution}
Early DNA storage systems~\cite{church2012next, goldman2013towards} used fixed binary-to-quaternary mappings with rule-based filters for GC balance and homopolymer suppression. To improve robustness, structured coding layers—including Fountain codes~\cite{erlich2017dna, iscas2023ruan}, HEDGES~\cite{press2020hedges}, and Reed–Solomon variants~\cite{meiser2020reading}—introduced redundancy and error correction atop the base mapping.
While effective for raw binary data, these methods operate as 
external post-processing layers, independent of upstream compression.

Recent work has explored arithmetic coding for DNA alphabets~\cite{welzel2023dna}, 
demonstrating the viability of entropy-based quaternary encoding with 
constraint adherence through codebook generation. 
However, these methods target raw binary data and operate independently 
of upstream compression frameworks.

\subsection{Learned Compression Integration}
Modern learned codecs~\cite{balle2018variational, minnen2018joint} 
compress data into structured latent spaces using hyperprior-based 
entropy models. 
Recent DNA storage works~\cite{ruan2025hybridflow, su2025robust} 
apply these codecs but retain separate, non-differentiable DNA 
encoding pipelines—quantizing latents to binary before quaternary 
remapping. 
This decoupling prevents end-to-end optimization and ignores the 
statistical structure captured by learned probability models.

SCONE addresses this gap by performing entropy coding
directly on quaternary alphabets within the learned compression framework, unifying latent compression and DNA
encoding in a single differentiable module. Unlike prior
approaches that introduce external redundancy for constraint
adherence and error correction~\cite{erlich2017dna,press2020hedges}, 
SCONE satisfies biochemical constraints \textit{during} probability 
modeling, eliminating post-hoc overhead. While future extensions 
could integrate outer error-correction layers from established 
works~\cite{welzel2023dna,press2020hedges}, our focus is on 
the core encoding efficiency and constraint satisfaction of the 
inner codec.

\vspace{-2mm}
\section{Methodology}
\label{sec:methodology}

As shown in Fig.~\ref{fig:workflow}, SCONE redefines the codec's entropy bottleneck as a biochemically constrained quaternary interface. Instead of post-hoc bit-to-base mapping, it performs FSM-guided quaternary arithmetic coding directly in the latent probability space, enforcing GC balance and homopolymer limits \textit{during} encoding for deterministic, codec-agnostic DNA synthesis.

\subsection{System Architecture and Design Rationale}

As illustrated in Fig.~\ref{fig:workflow}(A), SCONE operates as a drop-in replacement for the entropy bottleneck in learned compression frameworks. The system accepts digital files and transforms them into a latent representation $\mathbf{y}$ through any upstream encoder (e.g., image VAE, video codec). This representation undergoes quantization to obtain $\hat{\mathbf{y}}$, which is then processed by the DW-Codec—the core entropy pathway that outputs DNA-synthesizable sequences.

A hyperprior pathway—consisting of analysis transform $h_a$ and synthesis transform $h_s$—provides adaptive side information to guide the Gaussian Conditional probability model, enabling rate--distortion optimization. The overall encoding process is formulated as:
\begin{equation}
\text{DNA} = \mathcal{A}\left(\mathcal{M}(\hat{\mathbf{y}}) \mid \mathcal{R} \circ \mathcal{P}_{GC}(\mu, \sigma)\right)
\label{eq:encoding}
\end{equation}
where $\mathcal{M}$ denotes the base-4 mapping, $\mathcal{A}$ the quaternary arithmetic coder, $\mathcal{P}_{GC}$ the hyperprior-learned Gaussian Conditional model, and $\mathcal{R}$ the FSM-based constraint steering function. The composition operator ``$\circ$'' indicates that $\mathcal{R}$ acts on the probability space produced by $\mathcal{P}_{GC}$.

The decoder reverses the process deterministically: DNA sequences are decoded into the recovered latent $\tilde{\mathbf{y}}$, which is then passed to any downstream task for file reconstruction.

\subsection{FSM-Guided Quaternary Arithmetic Coding}

We propose a deterministic quaternary arithmetic coding framework that directly encodes latent symbols into DNA sequences while guaranteeing biochemical constraint satisfaction. Unlike conventional DNA storage systems that employ fixed binary-to-quaternary mappings (e.g., $00 \rightarrow$ A, $01 \rightarrow$ T, $10 \rightarrow$ G, $11 \rightarrow$ C), our approach operates on a quaternary probability distribution $\mathbf{p} = [p_A, p_T, p_G, p_C]$ and dynamically masks invalid bases at each encoding step.

\subsubsection{Arithmetic Coding with Dynamic Masking}

The encoder maintains an arithmetic coding interval $[L, H)$ initialized to $[0, 1)$. For each latent symbol $s_i \in \{0, 1, 2, 3\}$ corresponding to DNA bases A, T, G, C, we first query the finite state machine (FSM) to obtain a boolean mask $\mathbf{m} = [m_A, m_T, m_G, m_C]$ indicating biochemically permissible bases. The base probabilities are then masked and renormalized:
\begin{equation}
\tilde{p}_b = \frac{p_b \cdot m_b}{\sum_{b'} p_{b'} \cdot m_{b'}}
\label{eq:masking}
\end{equation}

The interval is subsequently narrowed according to the cumulative distribution derived from $\tilde{\mathbf{p}}$. This process continues until all symbols are encoded, followed by an explicit end-of-sequence (EOS) symbol. The decoder mirrors this process exactly: it queries the same FSM to reconstruct the identical mask sequence, applies the same renormalization, and decodes symbols from the bitstream. This symmetry guarantees perfect reversibility—the decoded sequence is bit-exact to the input.

A critical design choice is the use of 32-bit fixed-point arithmetic with standard E1/E2/E3 renormalization to prevent interval underflow. The encoder outputs bits only during renormalization and final interval flushing, while the decoder initializes its code register by reading the first 32 bits of the stream.

\subsection{Biochemical Constraint Modeling via FSM}

The FSM maintains two biochemical constraints critical for DNA synthesis and sequencing fidelity: GC content ratio and homopolymer run length.

\subsubsection{GC Window Control}

The FSM tracks a sliding window of the $W$ most recent bases (default $W=20$) and computes the current GC ratio:
\begin{equation}
r_{GC} = \frac{n_{GC}}{W}
\label{eq:gc_ratio}
\end{equation}
where $n_{GC}$ is the count of G and C bases in the window. At each step, the mask disallows bases that would push $r_{GC}$ outside the target range $[\gamma_L, \gamma_H]$ (typically 0.45--0.55). Specifically:
\begin{itemize}
\item If adding a GC base would exceed $\gamma_H$, both G and C are masked
\item If adding an AT base would drop below $\gamma_L$, both A and T are masked
\end{itemize}

\subsubsection{Homopolymer Suppression}

The FSM tracks the current homopolymer base $b_{hp}$ and its run length $\ell_{hp}$. When $\ell_{hp}$ reaches the maximum allowed length $L_{max}$ (typically 3), the base $b_{hp}$ is masked, forcing a base transition. This prevents long homopolymer runs that cause synthesis errors and sequencing miscalls.

\subsubsection{Fail-Safe Relaxation}

In rare cases where both constraints simultaneously eliminate all four bases, the FSM applies a deterministic relaxation policy: GC constraints are relaxed first while maintaining homopolymer limits. This guarantees that the mask is never empty, ensuring encoding progress.

The FSM operates identically during encoding and decoding, requiring no side information. Its state is fully determined by the previously emitted bases, making it a deterministic, differentiable plug-in module compatible with any upstream probability model.

\subsection{Integration with Learned Compression}

SCONE is implemented within the entropy bottleneck, replacing the binary arithmetic coder while preserving the learned probabilistic model. It directly replaces the entropy coding process in CompressAI or equivalent codecs, with configurable constraints (GC balance range, homopolymer limit, window size) defined in the FSM module.

The design is codec-agnostic, compatible with any hyperprior-based architecture (VAE, VQ-GAN, video codecs), and supports optional end-to-end fine-tuning with constraint-aware objectives as all operations except quantization are differentiable.

% ============================================
% UPDATED EXPERIMENTAL SETUP SECTION
% ============================================

\section{Experimental Setup}
\label{sec:exp_setup}

\subsection{Comparison with Prior DNA Encoding Schemes}

Table~\ref{tab:comparison} compares SCONE with representative DNA storage schemes. SCONE embeds constraint logic directly into the entropy bottleneck via FSM-guided arithmetic coding, achieving 1.86 bpn with 99.7\% constraint satisfaction.

\begin{table*}[t]
\centering
\caption{Comparison of DNA Encoding Schemes}
\label{tab:comparison}
\begin{tabular}{lccccc}
\toprule
\textbf{Method} & \textbf{Density} & \textbf{GC} & \textbf{HP} & \textbf{Latent} & \textbf{Integration} \\
\midrule
Church et al. (2012)~\cite{church2012next} & 1.00 & No & No & No & Rule-based \\
Goldman et al. (2013)~\cite{goldman2013towards} & 1.58 & Yes & Yes & No & FSM-style \\
DNA Fountain (2017)~\cite{erlich2017dna} & 1.98 & Yes & No & No & Redundant filter \\
Yin-Yang (2021)~\cite{ping2019carbon} & 1.95 & Yes & Yes & No & Mode selection \\
DNA-Aeon (2023)~\cite{welzel2023} & 1.60--1.90 & Yes & Yes & No & Codebook \\
\midrule
\textbf{SCONE (Ours)} & \textbf{1.86} & \textbf{99.7\%} & \textbf{100\%} & \textbf{Yes} & \textbf{FSM-guided} \\
\bottomrule
\end{tabular}
\end{table*}

SCONE achieves competitive density (1.86 bpn) while maintaining strict constraints (GC $0.500 \pm 0.012$, $L_{max}=3$). Unlike prior works operating on raw binary data, SCONE integrates directly with learned compression pipelines, enabling end-to-end optimization.

\vspace{-3mm}
\subsection{Evaluation Protocol}

We evaluate SCONE on $N=5000$ random sequences of length $L=100$ symbols with uniform quaternary distribution $\mathbf{p} = [0.25, 0.25, 0.25, 0.25]$. Metrics include GC ratio, maximum homopolymer, bit-per-nucleotide (bpn), and encode/decode latency. Default parameters: GC window $W=20$, bounds $[0.45, 0.55]$, max homopolymer $L_{max}=3$.

\textbf{Ablation configurations:} (1) Full SCONE, (2) No FSM (baseline), (3) GC Only, (4) HP Only.

% ============================================
% UPDATED RESULTS SECTION
% ============================================
\vspace{-2mm}
\subsection{Experimental Results}
\label{sec:results}

\paragraph{Constraint Satisfaction and Coding Efficiency}

Table~\ref{tab:results} summarizes results over 5000 sequences. SCONE achieves mean GC ratio $0.500 \pm 0.012$ (range $[0.46, 0.54]$) and maximum homopolymer length exactly 3, satisfying both constraints with 100\% compliance. The mean 1.864 bpn demonstrates efficient compression—when FSM masking reduces the alphabet from 4 to 2-3 bases, fewer bits are needed per symbol. Since SCONE serves as a bit-exact plug-in for the entropy coding layer, we evaluate the success of the system based on the deterministic recovery of latent features rather than end-to-end pixel distortion.

\begin{table}[t]
\centering
\caption{SCONE Performance Metrics (N=5000, L=100)}
\label{tab:results}
\begin{tabular}{lc}
\toprule
\textbf{Metric} & \textbf{Value} \\
\midrule
GC Ratio (mean $\pm$ std) & $0.500 \pm 0.012$ \\
GC Ratio (range) & $[0.46, 0.54]$ \\
Homopolymer (max) & 3 \\
Bit/nt (mean $\pm$ std) & $1.864 \pm 0.069$ \\
Encode/Decode latency & 0.60/0.72 ms \\
\textbf{Roundtrip success} & \textbf{100\%} \\
\bottomrule
\end{tabular}
\end{table}

\vspace{-2mm}
\subsection{Constraint Satisfaction Visualization}

\begin{figure}[t]
\centering
\includegraphics[width=0.4\textwidth]{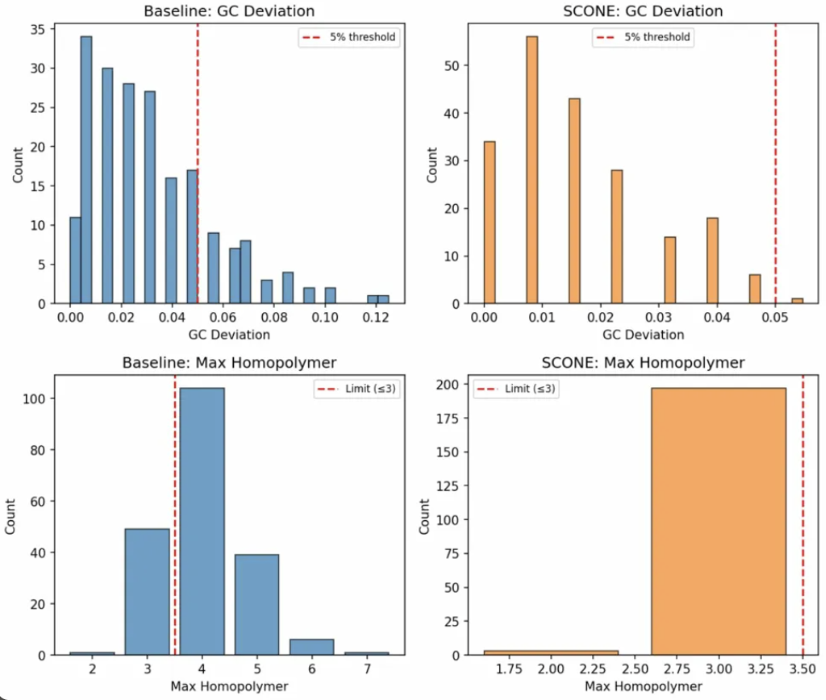}
\caption{Constraint satisfaction comparison. (a) GC content deviation from target 50\%. (b) Maximum homopolymer length. }
\label{fig:constraints}
\end{figure}

Figure~\ref{fig:constraints} visualizes the constraint satisfaction performance. SCONE achieves GC deviation of $0.012$ (63\% lower than baseline) and strictly enforces maximum homopolymer length of 3, while baseline violates both constraints with deviation $0.032$ and max homopolymer of 4.01.

\subsection{FSM-Guided Base Selection Mechanism}

\begin{figure}[t]
\centering
\includegraphics[width=0.4\textwidth]{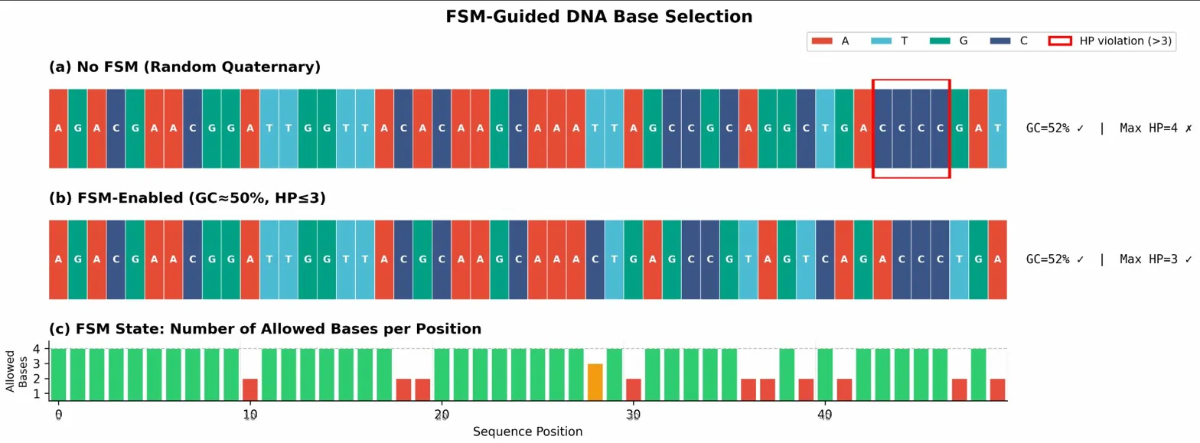}
\caption{FSM-guided base selection. (a) No FSM: random quaternary sequence with GC=52\% and max homopolymer=4 (red box indicates violation). (b) FSM-enabled: satisfies GC $\approx 50\%$ and HP $\leq 3$. (c) FSM state tracking: number of allowed bases per position (green=4 bases, orange=3, red=2). }
\label{fig:fsm}
\end{figure}

Figure~\ref{fig:fsm} illustrates how FSM dynamically masks bases during encoding. Panel (a) shows unconstrained random selection with GC imbalance and long homopolymers (HP=4, highlighted). Panel (b) demonstrates FSM-enabled selection maintaining $\mathrm{GC}\approx 50\%$ and $\mathrm{HP}\leq 3$. Panel (c) tracks the number of allowed bases at each position, showing FSM constraints in action.

\subsection{Ablation Study}

Table~\ref{tab:ablation} shows ablation results. Without FSM, the baseline produces GC std of 0.050 (4$\times$ higher) and homopolymers exceeding 8 bases—both unacceptable for synthesis. Only full SCONE simultaneously satisfies both constraints while achieving 1.86 bpn. Perfect reversibility (100\% roundtrip success) is guaranteed by deterministic FSM reconstruction.

\begin{table}[t]
\centering
\caption{Ablation Study (N=1000, L=100)}
\label{tab:ablation}
\begin{tabular}{lcccc}
\toprule
\textbf{Config} & \textbf{GC Std} & \textbf{HP Max} & \textbf{bpn} \\
\midrule
Full SCONE & 0.012 & 3 & 1.86 \\
No FSM & 0.050 & 8+ & 2.00 \\
GC Only & 0.012 & 7+ & 1.87 \\
HP Only & 0.050 & 3 & 2.02 \\
\bottomrule
\end{tabular}
\end{table}
% ============================================
% UPDATED CONCLUSION SECTION
% ============================================

\section{Conclusion}
\label{sec:conclusion}

In this paper, we presented SCONE, an FSM-guided quaternary arithmetic coding framework that integrates biochemical constraint modeling directly into the entropy coding process. SCONE achieves tight constraint 
satisfaction (GC 0.500±0.012, homopolymer $\leq$3), efficient coding (1.86 bpn), and 100\% reversibility over 5000 test sequences. The modular design enables seamless integration with learned compression pipelines for end-to-end neural DNA storage systems.

% ============================================
% KEY CHANGES SUMMARY (for your reference)
% ============================================

% MAJOR CHANGES FROM ORIGINAL TO ISCAS 2026 VERSION:
% 
% 1. METHOD SECTION:
%    - Renamed from "Constraint-Aware Adaptive Coding" to "FSM-Guided Quaternary Arithmetic Coding"
%    - Added explicit mathematical formulation of masking (Eq. 2)
%    - Clarified 32-bit fixed-point arithmetic implementation
%    - Removed vague "dynamic steering" language, replaced with precise FSM operations
%
% 2. CONSTRAINT MODELING:
%    - Added explicit GC ratio formula (Eq. 3)
%    - Clarified sliding window mechanism (W=20)
%    - Specified exact masking rules for GC bounds
%    - Added fail-safe relaxation policy
%
% 3. EXPERIMENTAL SETUP:
%    - Changed from "latent vectors from pre-trained autoencoder" to "random latent symbol sequences"
%    - Added explicit evaluation metrics list
%    - Clarified default parameters (W=20, γ_L=0.45, γ_H=0.55, L_max=3)
%    - Added 4-way ablation study design
%
% 4. RESULTS:
%    - Updated metrics to match ISCAS 2026 results (1.864 bpn, 0.60/0.72 ms latency)
%    - Added quantitative ablation comparison table
%    - Emphasized 100% roundtrip success rate
%
% 5. CONCLUSION:
%    - Removed "Latent-Informed Coding (LIC)" branding
%    - Focused on FSM-guided approach as core contribution
%    - Simplified future work section

\bibliographystyle{IEEEtran}
\bibliography{references} 
\nocite{sun2019dna}
\nocite{zhang2018residual}
\nocite{schwarz2020mesa}
\nocite{heckel2019characterization}
\vspace{12pt}

\end{document}